\documentclass[conference]{IEEEtran}
\IEEEoverridecommandlockouts
\usepackage{cite}
\usepackage{amsmath,amssymb,amsfonts}
\usepackage{algorithmic}
\usepackage{graphicx}
\usepackage{textcomp}
\usepackage{xcolor}

\usepackage{graphicx}
\usepackage{algorithm}
\usepackage{algorithmic}
\usepackage{amsthm}
\usepackage{amssymb}
\usepackage{multicol}
\usepackage{amsmath}

\usepackage{bbm,amssymb,amsmath,amsfonts,graphicx,fullpage,ifthen,twoopt,algorithm,algorithmic,color}

\usepackage{times}

\def\BibTeX{{\rm B\kern-.05em{\sc i\kern-.025em b}\kern-.08em
 T\kern-.1667em\lower.7ex\hbox{E}\kern-.125emX}}
\begin{document}

\title{Survey on Applications of Neurosymbolic Artificial Intelligence\\

}

\author{\IEEEauthorblockN{ Djallel Bouneffouf, Charu C. Aggarwal}
\IEEEauthorblockA{\textit{IBM Research}\\
Djallel.bouneffouf@ibm.com\\
New York, USA }
}

\maketitle

\begin{abstract}
In recent years, the Neurosymbolic framework has attracted a lot of attention in various applications, from recommender systems and information retrieval to healthcare and finance. This success is due to its stellar performance combined with attractive properties, such as learning and reasoning. The new emerging Neurosymbolic field is currently experiencing a renaissance, as novel frameworks and algorithms motivated by various practical applications are being introduced, building on top of the classical neural and reasoning problem setting. This article aims to provide a comprehensive review of significant recent developments in  real-world applications of Neurosymbolic Artificial Intelligence. Specifically, we introduce a taxonomy of common Neurosymbolic applications and summarize the state-of-the-art for each of those domains. Furthermore, we identify important current trends and provide new perspectives pertaining to the future of this burgeoning field.
\end{abstract}


\section{Introduction}
Neuro-Symbolic artificial intelligence focuses on bringing together the neural and the symbolic traditions in AI. Here, the term ``neural'' refers to the use of artificial neural networks and symbolic refers to the use of explicit symbol manipulation through formal logic. These fields have traditionally been separate, although recent years have seen the benefit of bringing them together in order to gain the best of both worlds.

Many practical applications require learning and reasoning abilities --- on the neural side, the desirable strengths would include learning from raw data, while on the symbolic side one would like to retain the natural explanatory power of these systems.  Examples of such applications include medical diagnosis~\cite{kang2021neuro}, recommender systems ~\cite{xian2020neural} and text-mining~\cite{kang2021neuro}, as well as the ease of making use of deep human expert knowledge in their design and function.

The drawbacks of neural networks lie in their inability to provide an explanation of the underlying reasoning mechanisms, which results in their being considered black-box models ~\cite{buhrmester2021analysis}. Symbolic systems can explain their inference process and use powerful declaration languages for knowledge representation. They allow explicit control, fast initial coding, dynamic variable binding, and knowledge abstraction ~\cite{velik2008neuro}. Problems of symbolic systems are their inability to handle incomplete information and to generalize. 

On the one hand, there are rule based systems, which require a lot of work in system development and adaptation. On the other hand there are supervised machine learning approaches, that require a lot of learning samples.

In terms of functional features, by utilizing symbolic approaches in combination with deep learning, one would hope to do better on issues like training from small data sets, recovery from errors, and in general, explainability, as opposed to systems that rely on deep learning alone.

We will now provide an extensive overview including various applications of the Neurosymbolic framework, both in real-life problem setting arising in multiple practical domains (healthcare, computer network routing, finance, and beyond), as well as in computer science and machine-learning in particular. 

\section{Real-Life Applications of Neurosymbolic }
As a general mathematical framework, the Neurosymbolic setting addresses the following challenges:

\section{Healthcare}
Artificial intelligence and machine learning methods offer a lot of promise in medicine.
In the health domain, the challenge is to build a transparent artificial intelligence. Among the fields which use AI techniques, is Drug Repurposing (DR) which involves finding a new indication for an existing drug. In this work\cite{drance2022neuro}, authors describe the use of neuro-symbolic algorithm in order to explain the process of link prediction in a knowledge graph-based computational drug repurposing. Link prediction consists of generating hypotheses about the relationships between a known molecule and a given target.

Author work done in \cite{lavin2021neuro} propose a novel probabilistic programmed deep kernel learning approach that works well in predicting cognitive decline in Alzheimer’s disease. 
The authors shows that the proposed method is a powerful tool in neurodegenerative disease diagnosis and prognostic modeling and they suggest that the system will perform well in other disease areas. 

The neuro-symbolic approach has advantageous medical-AI characteristics such as data-efficiency, model and prediction interpretability. At a higher level, they have compared two per spectives on AI-based longitudinal disease modeling: data driven with neural networks, and domain-engineered with probabilistic programs.

 \section{Image Processing}

Efficient multimedia event processing is a key enabler for real-time and complex decision
making in streaming media. The need for expressive queries to detect high-level human understandable spatial and temporal events in multimedia streams is inevitable due to the explosive growth of multimedia data in smart cities and internet \cite{bennetot2019towards}. The recent work in visual \cite{bennetot2019towards}, reasoning inspires the integration of visual and commonsense reasoning in multimedia processing. This work discuss why techniques integrating connectionist and symbolic paradigms are
the most efficient solutions to produce explanations and propose a reasoning model, to explain a neural network’s decision and use explanation in order to correct bias. 

\section{Business Management} 

The complexity of current organization systems, and the increase in importance of the realization of internal controls in firms, make it necessary to construct models that automate and facilitate the work of auditors. 

Authors in \cite{corchado2004neuro} developed a system to automate the internal control process. This system is composed of two case-based reasoning systems. The objective of the system is to facilitate the process of internal auditing in small and medium firms from the textile sector. The system, analyses the data that characterises each one of the activities carried out by the firm, then determines the state of each activity, calculates the associated risk, detects the erroneous processes, and generates recommendations to improve these processes. 

The system is a useful tool for the internal auditor in order to make decisions based on the risk generated. Each one of the case-based reasoning systems uses a different problem solving method in each of the steps of the reasoning cycle: fuzzy clustering during the retrieval phase, a radial basis function network and a multi-criterion discreet method during the reuse phase and a rule based system for recommendation generation. 

 \section{Brain and behavioral modeling.} 

Machine perception is a research field that is still in its infancy and is confronted  with many unsolved problems. In contrast, humans generally perceive their environment without problems. These facts were the motivation to develop a bionic model for human-like machine perception, which is based on neuroscientific and neuropsychological research findings about the  structural organization and function of the perceptual system of the human brain. Having systems available that are capable of a human-like perception of their environment would allow the automation of processes for which, today, human observers and their cognitive abilities are  necessary. The challenge to be faced is to merge and interpret large amounts of data coming from  different sources.

For this purpose, authors in \cite{velik2010towards} an information processing principle called neuro-symbolic  information processing is introduced using neuro-symbols as basic information processing units. Neuro-symbols are connected in a modular hierarchical fashion to a so-called neuro-symbolic  network to process sensor data. The architecture of the neuro-symbolic network is derived from  the structural organization of the perceptual system of the human brain. Connections and correlations between neuro-symbols can be acquired from examples in different learning phases. 
Besides sensor data processing, memory, knowledge, and focus of attention influence perception 
to resolve ambiguous sensory information and to devote processing power to relevant features.

 \section{Finance}

Assessment is the ability of dealing with loan demands. Different loan programs from different banks may be proposed according to the applicant’s characteristics \cite{hatzilygeroudis2011fuzzy}. 

For these tasks, authors in \cite{hatzilygeroudis2011fuzzy} present the design, implementation and evaluation of intelligent methods that assess bank loan applications where two separate intelligent systems have been developed and evaluated: a fuzzy expert system and a neuro-symbolic expert system. The former employs fuzzy rules based on knowledge elicited from experts. The latter is based on neurules, a type of neuro-symbolic rules that combine a symbolic and a connectionist representation. Neurules were produced from available patterns. 

Based on the results, we can say that synergies from using both (rule-based) domain theory and empirical data may result in effective systems.
 
 \section{Recommender Systems}

Recommender System (RS) is ubiquitous on today’s Internet to provide multifaceted personalized information services. While an enormous success has been made in pushing forward high-accuracy recommendations, the other side of the coin — the recommendation explainability — needs to be better handled for pursuing persuasiveness, especially for the era of deep learning based recommendation. 

In this paper,  the authors propose a model-based explainable recommendation approach, called  Neuro-Symbolic Interpretable Collaborative Filtering. the proposed system learns interpretable recommendation rules (consisting of user and item attributes) based on neural networks with two innovations: (1) a three-tower architecture tailored for the user and item sides in the RS domain; an improved architecture that is tailored for the user and item sides in recommendation; (2) fusing the powerful personalized representations of users and items to achieve adaptive rule weights and without sacrificing interpretability.  

Authors in \cite{Bouneffouf14,bouneffouf2012contextual} development of mobile applications provides a considerable amount of data of all types. In this sense, Mobile Context-aware Recommender Systems (MCRS) suggest the user suitable information depending on her/his situation and interests. the proposed work consists in applying machine learning  techniques and reasoning process in order to adapt dynamically the MCRS to the evolution of the user’s interest. To achieve this goal, the authors propose to combine bandit algorithm and case-based reasoning in order to define a contextual recommendation process based on different context dimensions (social, temporal and location). This paper describes a work on the implementation of 
a MCRS based on a $hybrid-\epsilon-greedy$ algorithm. It also presents results by comparing the $hybrid-\epsilon-greedy$ and the standard $\epsilon-greedy$ algorithm.

 \section{Natural language processing}
In recent years, AI research has demonstrated enormous potential for the benefit of humanity and society. 
While often better than its human counterparts in classification and pattern recognition tasks, however, AI still struggles with complex tasks that require commonsense reasoning such as natural language understanding. In this context, the key limitations of current AI models are: dependency, reproducibility, trustworthiness, interpretability, and explainability. In this work, \cite{cambria2022senticnet} they propose a commonsense framework that overcome these issues in sentiment analysis. They use auto-regressive language models to build trustworthy symbolic representations that convert natural language to a protolanguage and, extract polarity from text.As a result, SenticNet 7 is unsupervised, reproducible, interpretable, trustworthy, and explainable.

 \section{Information Retrieval}

 Authors in \cite{IRBouneffouf14,BouneffoufBG13a} development of mobile information retrieval. The proposed work consists in applying machine learning  techniques and reasoning process in order to adapt dynamically the information retrieval system to the evolution of the user’s interest. To achieve this goal, they propose to combine bandit algorithm and case-based reasoning in order to define a contextual information retrieval process based on different context dimensions (social, temporal and location). 
 
Part-of-speech tagging (POS) assigns grammatical tags (like noun, verb, etc.) to a word depending on its definition and its context. It is a first step before parsing may be applied. POS tagging has an important role in information retrieval. 
Authors in \cite{marques1aneuro} propose a new approach that conjoins both, by retaining the advantages of rule based methods and of machine learning methods. This is done by using a neural network but instead of randomizing the weights, they initialized the network using rules that express background knowledge.

They show that the combined system outperforms a machine learning system when only limited samples are available. Syntactic word-class tagging, usually assigns grammatical tags to a word, depending on its definition and its context. There are several applications of word tagging in text mining, either using the tags directly or by requiring the tags to become more semantically, or simply task oriented. 

They showed that the methods of neural symbolic integration can be success fully applied in the context of POS tagging. Furthermore, they showed that the embedding of generic background knowledge is particularly helpful if only limited training data is available. 

 \section{Language generation}
Although  they are currently riding a technological wave of personal assistants, many of these agents still struggle to communicate appropriately. 

Automated story generation is an area of AI research that aims to create agents that tell good stories. Previous story generation systems use symbolic representations to create new stories, but these systems require a vast amount of knowledge engineering. In contrast, neural language models lose coherence over time.

Authors in this work \cite{martin2021neurosymbolic} focus on the perceived coherence of stories. Where they created automated story generation systems that improved coherence by leveraging various symbolic approaches for neural systems. They did by separating semantic event generation from syntactic sentence generation, manipulating neural event generation to make it goal-driven, improving syntactic sentence generation to be more coherent, and creating a rule-based infrastructure to aid neural networks in causal reasoning.

 \section{Dialogue Systems}
 
The interpretability of task oriented dialogue systems is a big problem. Previously, most neural-based task-oriented dialogue systems employ an implicit reasoning
strategy that makes the model predictions uninterpretable to humans \cite{bouneffouf2021toward,bouneffouf2021toward1}. 

To obtain a trans parent reasoning process, authors in \cite{yang2022interpretable} introduce neuro symbolic to perform explicit reasoning that justifies model decisions by reasoning chains.
Since deriving reasoning chains requires multi hop reasoning for task-oriented dialogues, 
Authors propose a two phase approach that consists of a hypothesis generator and a reasoner. The system first obtain multiple potential operations to perform the desired task, through the hypothesis generator. Each hypothesis is then verified by the reasoner, and the valid one is selected to conduct the final prediction. 

\subsection{Question Answering} 

Understanding narratives requires reasoning about implicit world knowledge related to the causes, effects, and states of situations described in text. At the core of this challenge is how to access contextually relevant knowledge on demand and reason over it.
In this work \cite{bosselut2021dynamic},  authors present initial studies toward zero-shot commonsense question answering by formulating the task as inference over dynamically generated commonsense knowledge graphs. In contrast to previous studies for knowledge integration that rely on retrieval of existing knowledge from static knowledge graphs, the proposed system requires commonsense knowledge integration where contextually relevant knowledge is often not present in existing knowledge bases. Therefore, they 
present a novel approach that generates contextually-relevant symbolic knowledge structures on demand using generative neural commonsense knowledge models.
Empirical results on two datasets demonstrate the efficacy of the proposed approach for dynamically constructing knowledge graphs for reasoning. the proposed model approach achieves significant performance boosts over pretrained language models and vanilla knowledge models, all while providing interpretable reasoning paths for its predictions.

\section{Cyber Security}
 
A phishing attack is defined as a type of cybersecurity attack that uses URLs that lead to phishing sites and steals credentials and personal information. Since there is a limitation on traditional deep learning to detect phishing URLs from only the linguistic features of URLs, attempts have been made to detect the misclassified URLs by integrating security expert knowledge with deep learning \cite{park2021evolutionary}.
 
Authors in \cite{park2021evolutionary}, a genetic algorithm is proposed to find combinatorial optimization of logic programmed constraints and deep learning from given components, which are rule-based and neural component. The genetic algorithm explores numerous searching spaces of combinations of rules with deep learning to get an optimal combination of the components. 
 
\section{Education System}
 
Predicting student problem-solving strategies is a complex problem but one that can significantly impact automated instruction systems since they can adapt or personalize the system to suit the learner. While for small datasets, learning experts may be able to manually analyze data to infer student strategies, for large datasets, this approach is infeasible.
Authors in \cite{shakya2021student} develop a Machine Learning model to predict strategies from student data. While Deep Neural Network (DNN) based methods such as LSTMs can be applied for this task, they often have long convergence times for large datasets and like several other DNN-based methods have the inherent problem of overfitting the data. To address these issues, they develop a Neuro-symbolic approach for strategy prediction, namely a model that combines strengths of symbolic AI (that can encode domain knowledge) with DNNs. Specifically,  they encode relationships in the data using Markov Logic and use symmetries among these relationships to train an LSTM. 

\section{ Robotic}
In \cite{hanson2020neuro}, authors outline the design of novel robotic arms using convolutional neural networks and symbolic AI for logical control and affordance indexing. 
They describe robotic arms built with a humanlike mechanical configuration and aesthetic, with 28 degrees of freedom, touch sensors, and series elastic actuators. The arms were modeled with URDF models, and implemented motion control solutions for solving the casino card game, rock paper scissors, hand shaking, and drawing. The uses of the work extend across domains, and include arts and social human-robot interaction, as well as targeting more general co-bot applications. 

 \section{Smart City}
Smart building and smart city with innovative use cases are in need of Artificial 
Intelligence (AI). However, today’s AI mainly concerns machine learning, whereas the first years of AI were essentially about expert \cite{morel2021neuro}. 

Authors in \cite{morel2021neuro} propose to merge the two AI trends for a smart city, and point the way towards a complete integration of the two technologies, compatible with standard software.
The city and the building are complex systems, and developing new AI applications for urban systems  calls for the merging of a data-oriented approach (machine learning and ANN) with knowledge-based systems, and possibly other types of models.
With this objective, they argue that a Neuro-symbolic AI software architecture able to encapsulate all kinds of models and libraries should be based on (1) knowledge primitives at the conceptual level, (2) a knowledge representation language (resembling frames) for the symbolic level.

\section{Safe machine learning}
In the recent past, reinforcement learning has seen numerous advances and found applications in safety-critical settings \cite{BalakrishnanBMR19,zhao2022learning,epperlein2022reinforcement,nelson2022linearizing}. System failures in this setting can result in loss of property. Authors in \cite{anderson2020neurosymbolic}  takes a step towards solving this problem by guaranteeing that RL agents do not violate safety properties. they present REVEL, a partially neural reinforcement learning (RL) framework for provably safe exploration in continuous state and action spaces. The challenge here is to repeatedly verify neural networks within a learning loop without a high computational cost. The authors address this challenge using neurosymbolic approach with approximate gradients. At each iteration, it safely lifts a symbolic policy into the neurosymbolic space, performs safe gradient updates to the resulting policy, and projects the updated policy into the safe symbolic subset, all without requiring explicit verification of neural networks. 

\subsection{Neurosymbolic in Real-Life Applications: Summary and Analysis}
Their is different  ways to bring together the neural and symbolic traditions." In the brief address, only a few examples were given of corresponding work. In a much older survey from 2005 [4], work on NeSy AI was classified according to eight dimensions, grouped into three aspects:

\subsection{Interrelation}

(1) Integrated versus hybrid: here we are checking how the aggregation is done, hybrid or a symbolic  functionalities emerge from neural structures. 

(2) Neuronal versus connectionist: here the question is whether the artificial neural network architecture is chosen to mimic biological neural networks versus a more technological approach that focused on  computational properties.

(3) Local versus distributed: here we check how symbolic information is represented within the neural system, one dedicated neurons for a symbolic piece of knowledge) or a representation consists of a larger number of activation values.

(4) Standard versus nonstandard: whether the system implement a standard artificial neural network architectures that are widely used.
\begin{table}[ht]
\scriptsize
\caption {Neurosymbolic Interrelation for Each Domain}
\label{tab:Life0} 
\begin{tabular}{|l|r|l|l|l|}
\hline
                       & Integrated     & Neuronal         & Local            &  Standard    \\
                       &    vs          &  vs Connect-     &   vs Distri-     &  vs non- \\
                       &  Hybrid        &  ionist          &     buted        &  Standard \\
     \hline
Healthcare              &  H  &  C  &    L  &    S  \\ \hline
Image processing        &  H  &  N  &     L    &  S     \\ \hline
Business management     &  H  &  C  &    L     &   S    \\ \hline
Brain modeling          &  H  &  N  &    L     &   S    \\ \hline
Fiance                  &  H  &  C  &    L     &   S    \\ \hline
RS                      &  H  &  C  &    L     &   S    \\ \hline
NLP                     &  H  &  N  &    L     &   S     \\ \hline
Information Retrieval   &  H  &  C  &    L     &   S    \\ \hline
LG                      &  H  &  N   &    L     &  S     \\ \hline
Question answering      &  H  &  N   &    L     &  S     \\ \hline
Dialogue system         &  H  &  N   &    L     &  S     \\ \hline
Cyber security          &  H  &  C   &    L     &   S    \\ \hline
Education system        &  H  &  C   &    L     &   S    \\ \hline
Robotic                 &  H  &  C   &    L     &   S    \\ \hline
Smart city              &  H  &  C   &    L     &  S     \\ \hline
Safe machine learning   &  H  &  C   &    L     &  S     \\ \hline
\end{tabular}
\end{table}

from table \ref{tab:Life0} we can say that most Neurosymbolic technique are hybrid approach and standard, which mainly show a lack on the integrated technology that combine both learning and reasoning as one component, we also do not see much distributed work along the line of neural symbolic which could be a next challenge.
 
\subsection{Language}
- Symbolic versus logical: whether it is based on formal logic or structured datatypes like graphs.

- Propositional versus first-order: whether the underlying logic is propositional, or expressive ( quantifiers).

\begin{table}[ht]
\scriptsize
\caption {Neurosymbolic Language in each domain}
\label{tab:Life2} 
\begin{tabular}{|l|r|l|l|l|}
\hline
                  & symbolic & logical   & Propositional    & First-order \\\hline
Healthcare              &   $\surd$   &    &   $\surd$       &      \\ \hline
Image processing        &     &  $\surd$  &    $\surd$      &     \\ \hline
Business management     &     & $\surd$   &     $\surd$       &   \\ \hline
Brain modeling          &  $\surd$   &    &   $\surd$      &      \\ \hline
Finance                 &     &  $\surd$  &    $\surd$      &      \\ \hline
RS                 &     &  $\surd$  &    $\surd$      &      \\ \hline
NLP                     &  $\surd$    &    &   $\surd$       &       \\ \hline
LG                     &  $\surd$    &    &   $\surd$       &       \\ \hline
Information Retrieval  &  $\surd$    &    &  $\surd$        &      \\ \hline
Language generation     &     &  $\surd$  &         &   $\surd$   \\ \hline
Question answering      &  $\surd$     &    &    $\surd$       &      \\ \hline
Dialogue system         &     &  $\surd$    &         &  $\surd$      \\ \hline
Cyber security          &  $\surd$     &    &    $\surd$       &      \\ \hline
Education system        &     &    &         &      \\ \hline
Robotic                 &  $\surd$     &    &   $\surd$        &      \\ \hline
Smart city              &   $\surd$    &    &  $\surd$         &      \\ \hline
Safe machine learning   &     &    &         &   $\surd$    \\ \hline
\end{tabular}
\end{table}
 We can say from this table that the need of first order logic is not as high as a propositional logic. This could be explained by the simplicity of the propositional logic.
 
\subsection{Usage}
(7) Extraction versus representation: whether the information flow within the system primarily extracts symbolic information from a neural, or it is based on neural representations of symbolic knowledge.

(8) Learning versus reasoning: whether the system focus is on machine learning or on automated symbolic reasoning.
 
\begin{table}[ht]
\scriptsize
\caption {Neurosymbolic usage for each domain}
\label{tab:Life} 
\begin{tabular}{|l|r|l|l|l|}
\hline
     & Extraction  & Represent & Learning & Reasoning \\
     \hline
Healthcare              &     &  $\surd$  &   $\surd$      &      \\ \hline
Image processing        &     &  $\surd$  &   $\surd$      &      \\ \hline
Business management     &     &  $\surd$  &         &  $\surd$    \\ \hline
Brain modeling          &     &  $\surd$  &  $\surd$       &      \\ \hline
Fiance                  &  $\surd$   & $\surd$   &   $\surd$      &      \\ \hline
RS                      &     & $\surd$   &   $\surd$      &      \\ \hline
NLP                     &     & $\surd$   &   $\surd$      &       \\ \hline
Information Retrieval   &  $\surd$   & $\surd$   &   $\surd$      &      \\ \hline
Question answering      &     & $\surd$   &         &  $\surd$    \\ \hline
Dialogue system         &     &  $\surd$  &   $\surd$      &  $\surd$    \\ \hline
Cyber security          &     &  $\surd$  &  $\surd$        &      \\ \hline
Education system        &     &  $\surd$  &  $\surd$        &      \\ \hline
Robotic                 &     &  $\surd$  &  $\surd$        &      \\ \hline
Smart city              &     &  $\surd$  &         &  $\surd$     \\ \hline
Safe machine learning   &     &  $\surd$  &   $\surd$       &      \\ \hline
\end{tabular}
\end{table}

 We observe from this table that most applications are classification driven and representation driven. This observation gives an assessment of the actual need in term of reasoning vs classification in the different domains.  
 
 \section{Conclusions}
\label{sec:Conclusion}
In this article, we reviewed some of the most notable recent work on applications of  Neurosymbolic AI. We summarized, in an organized way (Tables 1, Tables 2 and Tables 3), various existing applications, by types of Neurosymbolic settings used, and discussed the advantages of using Neurosymbolic techniques in each domain. We briefly outlines of several important open problems and promising future extensions. 

In summary, the Neurosymbolic framework, is currently very active and promising research areas, and there are multiple novel techniques and applications emerging each year. We hope our survey can help the reader better understand some key aspects of this exciting field and get a better perspective on its notable advancements and future promises.

\bibliographystyle{ieeetr}
\bibliography{IEEEexample}

\begin{thebibliography}{10}

\bibitem{kang2021neuro}
T.~Kang, A.~Turfah, J.~Kim, A.~Perotte, and C.~Weng, ``A neuro-symbolic method
  for understanding free-text medical evidence,'' {\em Journal of the American
  Medical Informatics Association}, vol.~28, no.~8, pp.~1703--1711, 2021.

\bibitem{xian2020neural}
Y.~Xian, Z.~Fu, Q.~Huang, S.~Muthukrishnan, and Y.~Zhang, ``Neural-symbolic
  reasoning over knowledge graph for multi-stage explainable recommendation,''
  {\em arXiv preprint arXiv:2007.13207}, 2020.

\bibitem{buhrmester2021analysis}
V.~Buhrmester, D.~M{\"u}nch, and M.~Arens, ``Analysis of explainers of black
  box deep neural networks for computer vision: A survey,'' {\em Machine
  Learning and Knowledge Extraction}, vol.~3, no.~4, pp.~966--989, 2021.

\bibitem{velik2008neuro}
R.~Velik and D.~Bruckner, ``Neuro-symbolic networks: Introduction to a new
  information processing principle,'' in {\em 2008 6th IEEE International
  Conference on Industrial Informatics}, pp.~1042--1047, IEEE, 2008.

\bibitem{drance2022neuro}
M.~Dranc{\'e}, ``Neuro-symbolic xai: Application to drug repurposing for rare
  diseases,'' in {\em International Conference on Database Systems for Advanced
  Applications}, pp.~539--543, Springer, 2022.

\bibitem{lavin2021neuro}
A.~Lavin, ``Neuro-symbolic neurodegenerative disease modeling as probabilistic
  programmed deep kernels,'' in {\em International Workshop on Health
  Intelligence}, pp.~49--64, Springer, 2021.

\bibitem{bennetot2019towards}
A.~Bennetot, J.-L. Laurent, R.~Chatila, and N.~D{\'\i}az-Rodr{\'\i}guez,
  ``Towards explainable neural-symbolic visual reasoning,'' {\em arXiv preprint
  arXiv:1909.09065}, 2019.

\bibitem{corchado2004neuro}
J.~M. Corchado, M.~L. Borrajo, M.~A. Pellicer, and J.~C. Y{\'a}{\~n}ez,
  ``Neuro-symbolic system for business internal control,'' in {\em Industrial
  conference on data mining}, pp.~1--10, Springer, 2004.

\bibitem{velik2010towards}
R.~Velik, ``Towards human-like machine perception 2. 0,'' {\em International
  Review on Computers and Software}, vol.~5, no.~4, pp.~476--488, 2010.

\bibitem{hatzilygeroudis2011fuzzy}
I.~Hatzilygeroudis and J.~Prentzas, ``Fuzzy and neuro-symbolic approaches to
  assessment of bank loan applicants,'' in {\em Artificial Intelligence
  Applications and Innovations}, pp.~82--91, Springer, 2011.

\bibitem{Bouneffouf14}
D.~Bouneffouf, ``Freshness-aware thompson sampling,'' in {\em Neural
  Information Processing - 21st International Conference, {ICONIP} 2014,
  Kuching, Malaysia, November 3-6, 2014. Proceedings, Part {III}} (C.~K. Loo,
  K.~S. Yap, K.~W. Wong, A.~T.~B. Jin, and K.~Huang, eds.), vol.~8836 of {\em
  Lecture Notes in Computer Science}, pp.~373--380, Springer, 2014.

\bibitem{bouneffouf2012contextual}
D.~Bouneffouf, A.~Bouzeghoub, and A.~L. Gan{\c{c}}arski, ``A contextual-bandit
  algorithm for mobile context-aware recommender system,'' in {\em
  International Conference on Neural Information Processing}, pp.~324--331,
  Springer, 2012.

\bibitem{cambria2022senticnet}
E.~Cambria, Q.~Liu, S.~Decherchi, F.~Xing, and K.~Kwok, ``Senticnet 7: a
  commonsense-based neurosymbolic ai framework for explainable sentiment
  analysis,'' {\em Proceedings of LREC 2022}, 2022.

\bibitem{IRBouneffouf14}
D.~Bouneffouf, ``Context-based information retrieval in risky environment,''
  {\em Aust. J. Intell. Inf. Process. Syst.}, vol.~14, no.~1, 2014.

\bibitem{BouneffoufBG13a}
D.~Bouneffouf, A.~Bouzeghoub, and A.~L. Gan{\c{c}}arski, ``Contextual bandits
  for context-based information retrieval,'' in {\em Neural Information
  Processing - 20th International Conference, {ICONIP} 2013, Daegu, Korea,
  November 3-7, 2013. Proceedings, Part {II}} (M.~Lee, A.~Hirose, Z.~Hou, and
  R.~M. Kil, eds.), vol.~8227 of {\em Lecture Notes in Computer Science},
  pp.~35--42, Springer, 2013.

\bibitem{marques1aneuro}
N.~C. Marques1a, S.~Bader2b, V.~Rocio3c, and S.~H{\"o}lldobler2d,
  ``Neuro-symbolic word tagging,'' {\em International Review on Computers and
  Software}, 2021.

\bibitem{martin2021neurosymbolic}
L.~J. Martin, {\em Neurosymbolic Automated Story Generation}.
\newblock PhD thesis, Georgia Institute of Technology, 2021.

\bibitem{bouneffouf2021toward}
D.~Bouneffouf, R.~Feraud, S.~Upadhyay, M.~Agarwal, Y.~Khazaeni, and I.~Rish,
  ``Toward skills dialog orchestration with online learning,'' in {\em ICASSP
  2021-2021 IEEE International Conference on Acoustics, Speech and Signal
  Processing (ICASSP)}, pp.~3600--3604, IEEE, 2021.

\bibitem{bouneffouf2021toward1}
D.~Bouneffouf, R.~Feraud, S.~Upadhyay, I.~Rish, and Y.~Khazaeni, ``Toward
  optimal solution for the context-attentive bandit problem,'' in {\em
  International Joint Conference on Artificial Intelligence}, 2021.

\bibitem{yang2022interpretable}
S.~Yang, R.~Zhang, S.~Erfani, and J.~H. Lau, ``An interpretable neuro-symbolic
  reasoning framework for task-oriented dialogue generation,'' {\em arXiv
  preprint arXiv:2203.05843}, 2022.

\bibitem{bosselut2021dynamic}
A.~Bosselut, R.~Le~Bras, and Y.~Choi, ``Dynamic neuro-symbolic knowledge graph
  construction for zero-shot commonsense question answering,'' in {\em
  Proceedings of the 35th AAAI Conference on Artificial Intelligence (AAAI)},
  2021.

\bibitem{park2021evolutionary}
K.-W. Park, S.-J. Bu, and S.-B. Cho, ``Evolutionary optimization of
  neuro-symbolic integration for phishing url detection,'' in {\em
  International Conference on Hybrid Artificial Intelligence Systems},
  pp.~88--100, Springer, 2021.

\bibitem{shakya2021student}
A.~Shakya, V.~Rus, and D.~Venugopal, ``Student strategy prediction using a
  neuro-symbolic approach.,'' {\em International Educational Data Mining
  Society}, 2021.

\bibitem{hanson2020neuro}
D.~Hanson, A.~Imran, A.~Vellanki, and S.~Kanagaraj, ``A neuro-symbolic
  humanlike arm controller for sophia the robot,'' {\em arXiv preprint
  arXiv:2010.13983}, 2020.

\bibitem{morel2021neuro}
G.~Morel, ``Neuro-symbolic ai for the smart city,'' in {\em Journal of Physics:
  Conference Series}, p.~012018, IOP Publishing, 2021.

\bibitem{BalakrishnanBMR19}
A.~Balakrishnan, D.~Bouneffouf, N.~Mattei, and F.~Rossi, ``Incorporating
  behavioral constraints in online {AI} systems,'' in {\em The Thirty-Third
  {AAAI} Conference on Artificial Intelligence, {AAAI} 2019, The Thirty-First
  Innovative Applications of Artificial Intelligence Conference, {IAAI} 2019,
  The Ninth {AAAI} Symposium on Educational Advances in Artificial
  Intelligence, {EAAI} 2019, Honolulu, Hawaii, USA, January 27 - February 1,
  2019}, pp.~3--11, {AAAI} Press, 2019.

\bibitem{zhao2022learning}
P.~Zhao, P.~Ram, S.~Lu, Y.~Yao, D.~Bouneffouf, X.~Lin, and S.~Liu, ``Learning
  to generate image source-agnostic universal adversarial perturbations,'' in
  {\em International Joint Conference on Artificial Intelligence}, 2022.

\bibitem{epperlein2022reinforcement}
J.~P. Epperlein, R.~Overko, S.~Zhuk, C.~King, D.~Bouneffouf, A.~Cullen, and
  R.~Shorten, ``Reinforcement learning with algorithms from probabilistic
  structure estimation,'' {\em Automatica}, vol.~144, p.~110483, 2022.

\bibitem{nelson2022linearizing}
E.~Nelson, D.~Bhattacharjya, T.~Gao, M.~Liu, D.~Bouneffouf, and P.~Poupart,
  ``Linearizing contextual bandits with latent state dynamics,'' in {\em The
  38th Conference on Uncertainty in Artificial Intelligence}, 2022.

\bibitem{anderson2020neurosymbolic}
G.~Anderson, A.~Verma, I.~Dillig, and S.~Chaudhuri, ``Neurosymbolic
  reinforcement learning with formally verified exploration,'' {\em Advances in
  neural information processing systems}, vol.~33, pp.~6172--6183, 2020.

\end{thebibliography}

\end{document}